\documentclass[journal]{IEEEtran}
\IEEEoverridecommandlockouts
\usepackage{cite}
\usepackage{graphicx}
\usepackage{subcaption}
\usepackage{amsthm,amsmath,amssymb,amsfonts}
\usepackage{booktabs}
\usepackage{multirow}
\usepackage{graphicx}
\usepackage{textcomp}
\usepackage{xcolor}
\usepackage{mathrsfs}
\usepackage{amssymb}
\usepackage{pifont}

\newtheorem{definition}{Definition} 

\usepackage{textcomp}
\usepackage{bm}      
\usepackage{xcolor}
\usepackage{textcomp}
\usepackage{upgreek}
\usepackage{tabularx} 
\usepackage{arydshln}
\usepackage{cancel}
\usepackage{float}
\usepackage{array}  
\usepackage{multirow}  

\usepackage{makecell}   
\usepackage{booktabs}  
\usepackage{caption}    
\usepackage{siunitx}    

\usepackage{verbatim}
\allowdisplaybreaks[4]
\usepackage{graphicx}
\usepackage{float}

\usepackage{bm} 
\usepackage[ruled,linesnumbered]{algorithm2e}

\makeatletter
\renewcommand\normalsize{%
	\@setfontsize\normalsize\@xpt\@xiipt
	\abovedisplayskip 3\p@ \@plus5\p@ \@minus6\p@
	\abovedisplayshortskip \z@ \@plus5\p@
	\belowdisplayshortskip 5\p@ \@plus5\p@ \@minus5\p@
	\belowdisplayskip \abovedisplayskip
	\let\@listi\@listI}
\makeatother

\def\BibTeX{{\rm B\kern-.05em{\sc i\kern-.025em b}\kern-.08em
		T\kern-.1667em\lower.7ex\hbox{E}\kern-.125emX}}
\begin{document}

\title{Dynamic Multi-Agent Pickup and Delivery \\in Robotic Cellular Warehousing Systems}

\author{
Cheng Ren,~\IEEEmembership{Member,~IEEE},
Ming Li,~\IEEEmembership{Member,~IEEE},\\
Xinping Guan,~\IEEEmembership{Fellow,~IEEE},
and George Q. Huang,~\IEEEmembership{Fellow,~IEEE}%
\thanks{Corresponding author: George Q. Huang.}%
\thanks{The authors would like to express their sincere thanks to the financial support from the Research Grants Council of the Hong Kong Special Administrative Region, China (No.~C7076-22G, T32-707/22-N, 15215325, 15208824, 25208626, and T41-517/25-N) and an Innovation and Technology Funds from the Innovation and Technology Commission of the Hong Kong Special Administrative Region, China (No.~ITP/003/26LP).}%
\thanks{Cheng Ren, Ming Li, and George Q. Huang are with the Department of Industrial and Systems Engineering, The Hong Kong Polytechnic University, Hong Kong, China (e-mail: cheng.ren@polyu.edu.hk; ming.li@polyu.edu.hk; gq.huang@polyu.edu.hk).}%
\thanks{Xinping Guan is with the School of Automation and Intelligent Sensing, Shanghai Jiao Tong University, Shanghai, China (e-mail: xpguan@sjtu.edu.cn).}%
}

\maketitle

\begin{abstract}
Robotic cellular warehousing systems (RCWS) give rise to multi-agent pickup and delivery (MAPD) processes in which robots sequentially collect multiple stock-keeping units (SKUs) for each order. Unlike classical MAPD formulations that assume static tasks, real warehouse operations often involve dynamic order evolution, where new SKUs may be appended to an order while it is being executed. Motivated by this practical requirement, this letter formulates the Dynamic-MAPD problem considering internal order evolution for the first time.
Building on the token passing (TP) mechanism, we propose two event-triggered online replanning algorithms.
\textcolor{black}{The two strategies target different robot-resource configurations,
depending on whether additional robotic resources are available for cooperative assistance.}
The first, Dynamic-TP, enables an event-triggered dynamic response by allowing robots to replan from their current execution states through priority-aware token acquisition after order updates.
The second, Cooperative-TP, further enables reserved robots to assist newly added SKUs while preserving the original order ownership. Simulation results demonstrate that the proposed methods significantly reduce order flowtime compared with static and non-cooperative baselines, thereby improving the order fulfillment efficiency in RCWS.
\end{abstract}
\begin{IEEEkeywords}
Robotic warehouse, multi-agent pickup and delivery, order fulfillment, online replanning, dynamic order
\end{IEEEkeywords}

\section{Introduction}
Recently, robotic cellular warehousing systems (RCWS) have emerged as a promising architecture, consisting of multiple modular units known as RubikCells~\cite{ma2023rubikcell}. 
As shown in Fig. 1, when a robot receives an order containing multiple stock-keeping units (SKUs), it navigates to the corresponding storage modules, collects all required SKUs, and delivers them to the assigned packing station~\cite{ma2025operating}. 
This type of order-fulfillment mode in the RCWS can be naturally formulated as a multi-agent pickup and delivery (MAPD) process, 
which is an extension of the multi-agent path finding (MAPF) problem~\cite{stern2019multi}.

In classical MAPF settings, each agent is typically assigned a single start location and a single goal location, and the objective is to compute collision-free paths for all agents~\cite{zhao2024lexicographic}.
MAPD further integrates task assignment with path planning, requiring agents to pick up SKUs from the start location and deliver them to the designated destination~\cite{ma2017lifelong}.
Although existing MAPD methods can accommodate multiple pickup locations within an order, they generally assume that the set of pickup locations remains fixed throughout execution~\cite{salzman2020research}. However, unlike the static assumptions commonly made in previous work, real-world on-demand orders are inherently dynamic and uncertain.

\begin{figure}[!t]
    \centering
    \setlength{\belowcaptionskip}{0cm}
    \includegraphics[width=\linewidth]{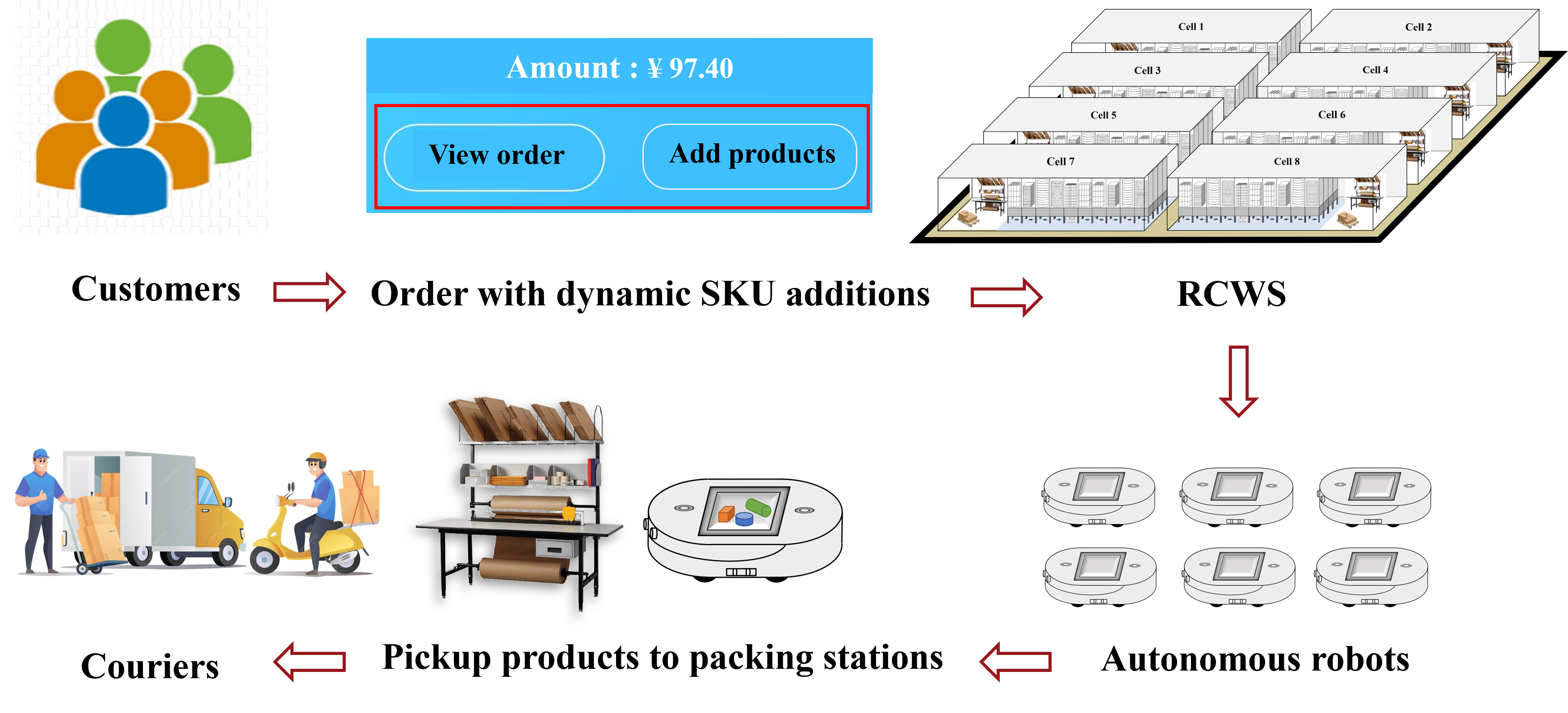}
    \caption{\textcolor{black}{Overview of order fulfillment process in one RCWS.}
    }
    \label{add}
\end{figure}

For example, Fig.~\ref{add} illustrates a typical user interface that allows customers to modify their orders before finalization~\cite{boysen2019warehousing}. These dynamic updates alter both the number and spatial distribution of SKUs to be retrieved, thereby invalidating the original pickup sequence of robots. Consequently, robots must promptly incorporate newly added SKUs into ongoing execution through route replanning and coordination while preserving feasibility and avoiding collisions. Otherwise, delayed incorporation of newly added SKUs may increase the completion time of the affected orders, thereby degrading overall order fulfillment efficiency.


This raises a fundamental question: how can the warehouse control system respond to uncertain SKU additions while maintaining efficient multi-robot coordination?
Accordingly, this letter focuses on a new form of MAPD in which ongoing orders may evolve during execution through the addition of new SKUs, rather than through the arrival of new tasks. Main contributions are summarized as follows.

\begin{enumerate}
   \item The \textbf{Dynamic-MAPD} problem is formalized for the first time by explicitly considering the endogenous uncertainty of orders. Already assigned orders may evolve during execution through the addition of new SKUs.

    \item A Dynamic Token Passing (Dynamic-TP) algorithm with event-triggered
    SKU-level path replanning is proposed. Newly added SKUs are integrated with
    the remaining uncollected SKUs, allowing the bound robot to replan its path
    from the current execution state.

    \item A Cooperative Token Passing (Cooperative-TP) algorithm is further
developed.  Newly added SKUs are selectively assigned to reserved robots
when cooperation is predicted to reduce the updated order's completion time, while preserving its primary robot ownership.
\end{enumerate}

The remainder of this letter is organized as follows. 
Section~II reviews related work. 
Section~III formulates the Dynamic-MAPD problem. 
Section~IV presents the proposed two algorithms. 
Section~V provides simulation results and performance analysis. 
Finally, Section~VI concludes the letter.

\section{Related Work}


Most follow-up research on MAPD has been motivated by robotic mobile fulfillment systems (RMFS), where robots operate under a goods-to-person (G2P) paradigm~\cite{zhao2024order}. 
Li \textit{et al.} extend MAPD to the double-deck RMFS architecture, coupling shelf relocation and robot navigation~\cite{li2023double}. Zhao \textit{et al.} propose a novel human–robot collaborative order-picking optimization problem in RMFS with multiple picking  stations and present a learning-based local search algorithm~\cite{zhao2025learning}.
Unlike the G2P picking mode in RMFS, researchers have generalized MAPD to robot-to-goods (R2G) like operations, which are naturally aligned with RCWS scenarios. Ma \textit{et al.} first formalize the MAPD problem and proposed the Token Passing (TP) and Token Passing with Task Swaps algorithms, which establish a benchmark for MAPD in R2G settings~\cite{ma2017lifelong}. 

Several MAPD variants further incorporate realistic constraints such as energy awareness, recharging, and distributed coordination. 
Bavaro \textit{et al.} integrate recharging decisions into the token-passing framework~\cite{bavaro2025multi}, 
and Camisa \textit{et al.} introduce a distributed MAPD formulation that achieves scalability via primal decomposition~\cite{camisa2022multi}. 
More recent formulations introduce practical considerations such as task deadlines~\cite{makino2024online}, external agents sharing the same environment~\cite{bonalumi2025multi}, and execution delays~\cite{lodigiani2023robust}. 
\textcolor{black}{Makino \textit{et al.} introduce the MAPD with task deadlines, where each order is associated with a deadline. They propose deadline-aware token passing and its task-swapping variant Dynamic-TPTS to reduce tardiness in online environments~\cite{makino2024online}. 
Bonalumi \textit{et al.} study MAPD with external agents, where a team of robots must accomplish delivery tasks while sharing the environment with independent external agents whose behaviors are unknown, and proposed modeling-based approaches to anticipate conflicts and plan collision-free paths~\cite{bonalumi2025multi}. 
Lodigiani \textit{et al.} address MAPD with delays, where robots may not perfectly follow their planned paths due to runtime disturbances~\cite{lodigiani2023robust}.}


Furthermore, multi-goal MAPD extends each task to a sequence of destinations, representing multi-SKU orders that must be picked before delivery~\cite{xu2022multi,zhong2022optimal} . 
Xu \textit{et al.} design a large neighborhood search strategy to scale multi-goal MAPD to large warehouses~\cite{xu2022multi}, 
and Zhong \textit{et al.} analyze its theoretical complexity and provided optimality bounds~\cite{zhong2022optimal}. Kudo \textit{et al.} extend the conventional single-task MAPD formulation to a multi-task setting, where each agent can be assigned multiple tasks simultaneously under payload constraints~\cite{kudo2023tsp}. 

Dynamic pickup and delivery have been widely studied in the vehicle routing problem (VRP) literature~\cite{cai2023survey}. While both VRP and MAPD are defined on graphs, they differ in key aspects. 
VRP typically assumes independent vehicles on road networks without explicit collision avoidance, whereas MAPD involves tightly coupled multi-robot coordination in a shared workspace with spatial–temporal conflicts. 
Moreover, VRP is defined on sparse road networks, while warehouse environments are dense grid-like graphs with frequent interactions. 
Therefore, VRP-based methods cannot be directly applied without incorporating collision-aware multi-agent planning.

To the best of our knowledge, prior MAPD studies have not addressed execution-time SKU additions to ongoing orders. Conventional TP assumes fixed tasks and does not handle order changes during execution~\cite{ma2017lifelong}. This work fills this gap by formulating Dynamic-MAPD and proposing an event-triggered TP framework for evolving orders.

\begin{table}[!t]  
    \centering  
    \small 
    \caption{Comparison of MAPD Variants and This Work} 
    \label{tab:comparison}
    \renewcommand\arraystretch{1}
    \resizebox{\linewidth}{!}{
    \begin{tabular}{c|c|c} 
    \hline 
    Reference & Order dynamics & Research focus\\ 
    \hline 
   \cite{li2023double}         & Static & Double-deck MAPD with shelf relocation\\
   \cite{bavaro2025multi}      & Static & Battery-aware MAPD with recharge scheduling\\
    \cite{camisa2022multi}      & Static & Distributed MAPD with scalable decomposition\\
       \cite{makino2024online}      & Static & Task deadline-aware MAPD \\
       \cite{bonalumi2025multi}     & Static & MAPD with external agents\\
       \cite{lodigiani2023robust}      & Static &  MAPD with execution delays\\
        \cite{xu2022multi}          & Static & Multi-goal MAPD with large neighborhood search\\
    \cite{zhong2022optimal}     & Static & Theoretical analysis of multi-goal MAPD\\ 
   \cite{kudo2023tsp}          & Static & Multi-order planning with TSP-based routing\\ 
        \hline 
    \textbf{Ours}               & \textbf{Dynamic} & \textbf{Dynamic orders and event-driven online response}\\ 
    \hline 
    \end{tabular}}
\end{table}

\section{Dynamic MAPD Problem Formulation}
\begin{figure}[!t]
    \centering
    \setlength{\belowcaptionskip}{0cm}
    \includegraphics[width=\linewidth]{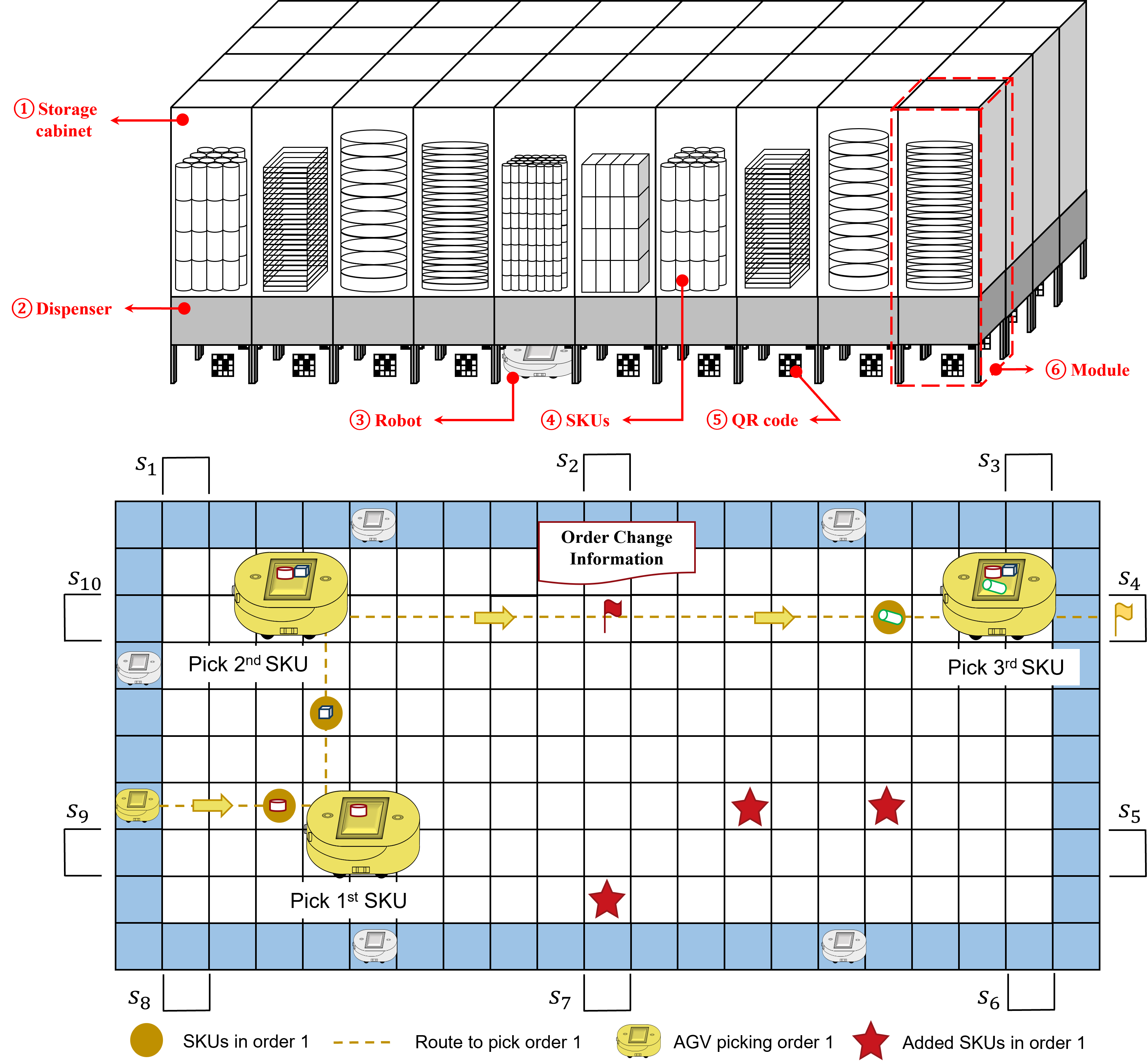}
    \caption{RubikCell and an order with dynamic additions.}
    \label{RCWS}
 \end{figure}

Fig.~\ref{RCWS} displays a schematic diagram of the RubikCell. 
Each RubikCell is represented by an undirected grid graph $G=(V,E)$, 
where each vertex $v=(x,y) \in V$ corresponds to a storage cabinet, 
and each edge $e \in E$ connects two adjacent vertices satisfying 
$|x-x'|+|y-y'|=1$, enforcing four-directional (north, east, south, west) movement 
while excluding diagonal connections. \textcolor{black}{The storage area of a RubikCell is modeled as a rectangular grid of size 
$N=L_x \times L_y$, where $L_x$ and $L_y$ denote the numbers of storage cabinets 
along the horizontal and vertical directions, respectively.} 
Let $\mathcal{M}=\{m_1,\dots,m_J\}$ denote the set of storage modules, 
where module $m_j$ is located at vertex $v_{m_j}\in V$ and stores one product type.
A set of $I$ packing stations $\mathcal{S}=\{S_1,\dots,S_I\}$ are placed along the 
boundaries for order delivery.

Multiple robots operate beneath modular storage dispensers to fulfill multi-SKU orders. The robot set is $\mathcal{A}=\{a_1,\dots,a_n\}$, 
where robot $a_i$ has an initial position $p_i(0)\in V$.
Since each robot completes its assigned order by delivering the picked SKUs to a packing station,
we assume that each robot is initially located at a packing station, i.e.,$ p_i(0)\in \mathcal{S}, \forall a_i\in \mathcal{A}.$
At discrete time $t$, the position of robot $a_i$ is $p_i(t)$, 
and its executed path is denoted by 
$\mathcal{P}_i(t)=\{p_i(0),p_i(1),\dots,p_i(t)\}$.

Let $\mathcal{O}=\{O_1,\dots,O_K\}$ denote the set of all orders released to the system. 
Each order $O_k\in\mathcal{O}$ is represented as the set of storage locations corresponding to its required SKUs:
\begin{equation}
    O_k = \{p_{k,1}, p_{k,2}, \dots, p_{k,n_k}\},
\end{equation}
where $n_k$ is the number of products initially included in order $O_k$.
To capture the dynamic nature of the warehouse, each order can receive 
additional SKUs during execution.  
Let $\gamma_k \in \{0,1\}$ indicate whether order $O_k$ receives a dynamic SKU update during its execution, where
$\gamma_k \sim \mathrm{Bernoulli}(p_{\mathrm{add},k})$ and
$p_{\mathrm{add},k}\in [0,1]$ denotes the probability that order $O_k$ is augmented with additional SKUs during its execution.
If $\gamma_k=1$, let $t_k^{\mathrm{upd}}$ denote the time at which the dynamic update occurs.
Conditional on $\gamma_k=1$, let $n_k^{\mathrm{add}}$ denote the number of newly added SKUs, illustrated by the red five-pointed stars in Fig.~2. Their storage locations $\{u_{k,1},\dots,u_{k,n_k^{\mathrm{add}}}\}$ 
are sampled i.i.d. from a spatial distribution $\pi_k(\cdot)$ defined on $V$.
The newly added SKUs at update time $t_k^{\mathrm{upd}}$ are denoted by
\begin{equation}
    \Delta O_k(t_k^{\mathrm{upd}})=\{u_{k,1},\dots,u_{k,n_k^{\mathrm{add}}}\},
\end{equation}
and the updated order becomes
\begin{equation}
   O_k^{\mathrm{new}}(t_k^{\mathrm{upd}})
   =O_k^{\mathrm{rem}}\cup\Delta O_k(t_k^{\mathrm{upd}}),
\end{equation}
where $O_k^{\mathrm{rem}}$ is the remaining uncollected SKUs before the update. In practical warehouse operations, different orders may have distinct delivery deadlines according to their service types.
For instance, some customer orders correspond to instant delivery, requiring immediate fulfillment, while others represent scheduled deliveries that can be dispatched later.
Accordingly, each order $O_k$ is associated with a delivery deadline $d_k$, representing its required completion time.

\textbf{Assumption 1 (Perfect synchronization).} 
The system controller is instantly notified of dynamic order updates and synchronizes the corresponding order information to the shared token without communication delay or data loss.

\textbf{Assumption 2 (Single Update per Order).} 
Each order is allowed to receive at most one dynamic SKU addition during its execution. 
Once the order update is processed, no further modifications are permitted until the order is completed.

\textcolor{black}{Although Assumption 2 considers at most one update per order for clarity, the proposed framework can be naturally extended to multiple updates. Each new SKU addition can be treated as an independent dynamic event, and the event-triggered coordination mechanism can be applied iteratively.}

\begin{definition}  (\textbf{Dynamic-MAPD Problem}).
\textit{The Dynamic Multi-Agent Pickup and Delivery problem is defined on a graph-based multi-agent system where each order $O_k$ consists of a set of goal locations to be visited before delivery. 
During execution, the goal set of $O_k$ may dynamically change, e.g., through the addition of new goals. 
The objective is to minimize the total order flowtime while ensuring motion feasibility, collision avoidance, and delivery deadlines.}
\end{definition}

In the RCWS, these goal locations correspond to SKU storage locations within a RubikCell, and $\Delta O_k(t)$ represents newly added SKUs during order execution.
Let $\pi$ denote a coordination policy that determines task assignment and path replanning decisions.
Let $T_k$ denote the completion time of order $O_k$, defined as the time when all SKUs associated with $O_k$, including any dynamically added ones, have been picked and delivered to the designated packing station. 
The Dynamic-MAPD problem aims to minimize the total order flowtime, which is equivalently the sum of order completion times:
\begin{equation}
    \min_{\pi} \sum_{k=1}^{K} T_k,
\end{equation}
subject to the following constraints.

\begin{equation}
\begin{aligned}
\text{(C1)}\;& p_i(t{+}1) \in \mathcal{N}(p_i(t)), \quad \forall i,t, \\[2pt]
\text{(C2)}\;& p_i(t) \neq p_j(t), \quad \forall i \neq j,\forall t, \\[2pt]
\text{(C3)}\;& (p_i(t),p_i(t{+}1)) \neq (p_j(t{+}1),p_j(t)),
\quad \forall i \neq j,\forall t, \\[2pt]
\text{(C4)}\;& T_k \le d_k, \quad \forall k .
\end{aligned}\nonumber
\end{equation}

Constraints (C1)-(C3) ensure discrete motion feasibility and collision avoidance, including both vertex conflicts and edge swap conflicts, while (C4) enforces order deadline constraints.

\section{Event-Triggered Token Passing}
In this section, we extend the classical TP scheme to address the proposed Dynamic-MAPD problem by introducing an event-triggered token-access mechanism.
At $t=0$, the initially released orders are assigned to 
available robots using the conventional TP mechanism~\cite{ma2017lifelong}. 
Each robot acquires the token, selects an unassigned order, and computes 
a collision-free pickup-and-delivery route. The resulting order-robot 
assignments and reserved routes are stored in the shared token, providing 
the initial execution plan.

During execution, the addition of new SKUs to ongoing orders triggers 
the proposed dynamic response mechanism. If multiple orders are updated 
simultaneously, they access the token in ascending order of their 
remaining time to deadline, defined as 
$\Delta d_k(t)=d_k-t$. After obtaining the token, the bound robot merges 
the newly added SKUs with the remaining uncollected SKUs of its original 
order and replans a collision-free route from its current state 
to collect these SKUs and reach a packing station, while respecting the 
routes reserved in the token. The updated path is then stored in the 
token and executed directly without invoking an additional round of 
conventional TP. \textcolor{black}{This event-triggered procedure enables newly added SKUs to be incorporated timely into ongoing order execution rather than being deferred until the
original order is completed. Specifically, two execution modes are developed
for different robot-resource configurations. The binding mode handles each
updated order using its originally assigned robot without requiring additional
cooperative resources, whereas the cooperative mode uses reserved robots to
reduce order completion time through parallel execution.}


\begin{definition}[Binding Mode]
Each order $O_k$ is permanently bound to one robot $a_i$ once execution begins.  
Any newly added items $\Delta O_k(t)$ must be collected by the same robot.
\end{definition}

\begin{definition}[Cooperative Mode]
Additional robots are reserved for cooperative assistance. When an update
$\Delta O_k(t)$ occurs, a reserved robot may be assigned to collect the newly
added SKUs while the primary robot continues serving the remaining original
SKUs. All SKUs are ultimately delivered to the same packing station.
\end{definition}

\begin{algorithm}[!t]
\caption{Dynamic Token Passing}
\label{alg:dtp}
\LinesNumbered
\SetAlgoLined
\KwIn{$G=(V,E)$; robots $\mathcal{A}$; stations $\mathcal{S}$; token $\mathcal{T}$; deadlines $\{d_k\}$}
\KwOut{Updated reserved paths $\{\mathcal{P}_i^{\mathrm{res}}\}$ stored in $\mathcal{T}$}

Initialize token $\mathcal{T}$ with initial assignments and routes\;
  \textcolor{black}{Initialize the unfinished-order index set
$\mathcal U\leftarrow\emptyset$\;}

\While{there exists unfinished orders}{
    Monitor indicators $\{\gamma_k(t)\}$ for all orders $\{O_k\}$\;
    Construct the updated-order index set
$\mathcal{K}_{\mathrm{upd}}(t)
\triangleq\{k\mid\gamma_k(t)=1\}$\;

  \If{ \textcolor{black}{$\mathcal{K}_{\mathrm{upd}}(t)\cup\mathcal{U}
\neq\emptyset$}}{

    \ForEach{$k\in\mathcal{K}_{\mathrm{upd}}(t)$}{
        Merge order items:
        $O_k^{\mathrm{new}}(t)
        \leftarrow O_k^{\mathrm{rem}}\cup\Delta O_k(t)$\;
    }

    \ForEach{ \textcolor{black}{$k\in
    \mathcal{K}_{\mathrm{upd}}(t)\cup\mathcal{U}$}}{
        Compute the remaining time to deadline:
        $\Delta d_k(t)\leftarrow d_k-t$\;
    }

      \textcolor{black}{Build a priority queue $\mathcal{Q}$ from
    $\mathcal{K}_{\mathrm{upd}}(t)\cup\mathcal{U}$,
    sorted by ascending $\Delta d_k(t)$\;}

        \While{$\mathcal{Q}\neq\emptyset$}{
   Select the most urgent order index 
$k^\star\leftarrow\arg\min_{k\in\mathcal Q}\Delta d_k(t)$;
            Let $a_{i^\star}\leftarrow a(k^\star)$ be its bound robot\;

            Robot $a_{i^\star}$ acquires token $\mathcal{T}$\;
            Attempt to replan from current position to complete $O_{k^\star}^{\mathrm{new}}(t)$ w.r.t.\ reserved paths in $\mathcal{T}$ using $\texttt{Path1}(\cdot)$\;

            \If{$\texttt{Path1}(\cdot)$ succeeds}{
                Commit the resulting complete reserved path $\mathcal{P}_{i^\star}^{\mathrm{res}}$ into $\mathcal{T}$\;
                 \textcolor{black}{Remove $k^\star$ from $\mathcal U$ if present\;}
            }
            \Else{
                Apply $\texttt{Path2}(\cdot)$ to generate a feasible partial path to a safe endpoint\;
                Commit the partial reserved path $\mathcal{P}_{i^\star}^{\mathrm{res}}$ into $\mathcal{T}$\;
               \textcolor{black}{Add $k^\star$ to $\mathcal U$ for future replanning if not already present\;}
            }

            Release token $\mathcal{T}$\;
            Remove $k^\star$ from the priority queue $\mathcal{Q}$\;
        }
    }
      \textcolor{black}{Robots advance one time step along their latest reserved paths;}
}
\end{algorithm}

\subsection{Dynamic Token Passing}
Dynamic-TP handles dynamically evolving orders under a binding policy through an event-triggered token-access mechanism.
The system operates over an unbounded execution horizon starting from $t=0$ until all assigned orders are completed.
Initially, the shared token $\mathcal{T}$ is constructed to store the initial task assignments and reserved paths of all robots (Line~1).
Robots then execute their reserved paths synchronously, advancing step by step according to the paths.

When an order update occurs, the robot selected to acquire the token immediately performs local replanning from its current execution state.
The remaining products of its original order are merged with the newly added products, and a new collision-free path to a packing station is computed with respect to the paths stored in the token.
The updated path is then written back to the token, ensuring global consistency.

\textcolor{black}{Let $\gamma_k(t)=1$ indicate that the update of order $O_k$ occurs at
time $t$, and $\gamma_k(t)=0$ otherwise. At each time step, Dynamic-TP monitors the order update indicators
$\gamma_k(t)$ of all orders (Line~4). If no new update is detected and the
unfinished-order set $\mathcal{U}$ is empty, robots continue executing their
current reserved paths without replanning. Otherwise, newly updated orders and
previously unfinished orders in $\mathcal{U}$ are included in the same priority
queue for token access and replanning.}
Based on the detected update events, the updated-order index set
$\mathcal{K}_{\mathrm{upd}}(t)$ is constructed (Line~5). For each newly updated
order $k\in\mathcal{K}_{\mathrm{upd}}(t)$, the newly added SKUs
$\Delta O_k(t)$ are merged with the remaining uncollected SKUs
$O_k^{\mathrm{rem}}$ to form the updated order
$O_k^{\mathrm{new}}(t)$ (Lines~7--9). The remaining time to deadline
$\Delta d_k(t)$ is then computed for all orders in
$\mathcal{K}_{\mathrm{upd}}(t)\cup\mathcal{U}$ to quantify their replanning
urgency (Lines~10--12). The priority queue $\mathcal{Q}$ is subsequently
constructed from $\mathcal{K}_{\mathrm{upd}}(t)\cup\mathcal{U}$ and sorted in
ascending order of $\Delta d_k(t)$ (Line~13). Consequently, orders closer to
their deadlines are processed first, while unfinished orders retained in
$\mathcal{U}$ can re-enter the replanning procedure even when no new update
event is detected at the current time step.

Dynamic-TP then processes the queue sequentially. At each iteration, the most urgent order $k^\star$ is selected, and its bound robot $a_{i^\star}$ is identified and acquires the token (Lines~14-16). The robot then attempts to replan from its current position to complete the updated order $O_{k^\star}^{\mathrm{new}}(t)$ with respect to the reserved paths stored in the token using $\texttt{Path1}(\cdot)$ (Line~17). \textcolor{black}{
Function $\texttt{Path1}(\cdot)$ first constructs the visiting sequence
of the remaining pickup locations using a nearest-neighbor heuristic.
Starting from the robot's current position, it iteratively selects the
nearest unvisited pickup location, and the packing station is appended
as the final destination. Given this visiting sequence, the underlying
dynamic A* planner is invoked once on the time-expanded graph to compute
a complete path that visits the pickup locations in the determined order
and then reaches the packing station. During this planning call, the
spatiotemporal reservations stored in the token are treated as dynamic
obstacles, thereby ensuring that the path is collision-free.
}

  \textcolor{black}{If $\texttt{Path1}(\cdot)$ succeeds, the resulting complete reserved path
$\mathcal{P}_{i^\star}^{\mathrm{res}}$ is committed to the token, and
$k^\star$ is removed from the unfinished-order index set $\mathcal{U}$,
if present (Lines~18--21). Otherwise, $\texttt{Path2}(\cdot)$ is applied
to generate a feasible partial path to a safe endpoint, and the resulting
partial reserved path $\mathcal{P}_{i^\star}^{\mathrm{res}}$ is committed
to the token (Lines~22--24). Function $\texttt{Path2}(\cdot)$ similarly
employs dynamic A* search on the time-expanded graph to compute a
collision-free relocation path from the robot's current position to a
designated safe endpoint that does not conflict with existing
spatiotemporal reservations or active and potential task locations. In
this case, $k^\star$ is added to $\mathcal{U}$, if not already present,
for replanning at a subsequent time step (Line~25). The token is then
released (Line~27), and $k^\star$ is removed from the current priority
queue $\mathcal{Q}$ (Line~28).
After all orders in the current priority queue have been processed, all robots simultaneously advance one time step along their latest reserved paths stored in the token (Line~31). Therefore, update-triggered replanning and path execution are explicitly separated within each discrete time step.}

\subsection{Cooperative Token Passing}
\begin{algorithm}[!t]
\caption{Cooperative Token Passing}
\label{alg:dtpcoop}
\LinesNumbered
\SetAlgoLined
\KwIn{$G=(V,E)$; robots $\mathcal{A}$; stations $\mathcal{S}$; token $\mathcal{T}$}
\KwOut{Updated reserved paths $\{\mathcal{P}_i^{\mathrm{res}}\}$ stored in $\mathcal{T}$}

\ForEach{updated order $O_k$ at time $t_{\mathrm{upd}}$}{
   Identify bound robot $a_i$ and reserved robot set $\mathcal{A}_{\mathrm{res}}$;

    Compute tentative baseline path $P_i^{\mathrm{base}}$ to complete $O_k^{\mathrm{rem}}\cup\Delta O_k(t_{\mathrm{upd}})$\;
    Compute $\mathrm{ETA}_i^{\mathrm{base}}$\;

    Compute tentative remaining-task path $P_i^{\mathrm{rem}}$ to complete $O_k^{\mathrm{rem}}$\;
    Compute $\mathrm{ETA}_i^{\mathrm{rem}}$\;

    \ForEach{$a_j\in\mathcal{A}_{\mathrm{res}}$}{
        Compute tentative add-pickup path $P_j^{\mathrm{add}}$ for $\Delta O_k(t_{\mathrm{upd}})$ to station $S_k$\;
        Compute $\mathrm{ETA}_j^{\mathrm{add}}$\;
    }

    \If{$\exists a_{j^\star}\in\mathcal{A}_{\mathrm{res}}$ satisfying
    $\max\left(\mathrm{ETA}_{j^\star}^{\mathrm{add}}, \mathrm{ETA}_i^{\mathrm{rem}}\right)
< \mathrm{ETA}_i^{\mathrm{base}}$}{
        Assign $\Delta O_k(t_{\mathrm{upd}})$ to $a_{j^\star}$\;
        Remove added SKUs from the pickup points of $a_i$ in token $\mathcal{T}$\;
        Insert $a_{j^\star}$ into the replanning set and plan paths for the assisted robot $a_{j^\star}$  using $\texttt{Path1}(\cdot)$\; 
    }
}

\end{algorithm}

\textcolor{black}{While Dynamic-TP enforces a binding mode in which each order is served by a single robot, Cooperative-TP allows reserved robots to assist dynamically updated orders. Newly added SKUs may be reassigned to a reserved robot while the primary robot continues serving the remaining original SKUs. This can reduce order flowtime through parallel execution, but at the cost of occupying additional robotic resources.}

\textcolor{black}{Cooperation is activated only when it is predicted to outperform the baseline Dynamic-TP execution in terms of order completion time. Inspired by the task-swap mechanism in~\cite{ma2017lifelong}, Cooperative-TP evaluates the overall completion time under cooperative and non-cooperative execution rather than relying on a single pickup-time comparison.}

The complete procedure of Cooperative-TP is summarized in Algorithm~2. 
When an order update $\Delta O_k(t_{\mathrm{upd}})$ occurs, Cooperative-TP first identifies the bound robot $a_i$ executing order $O_k$, as well as the set of reserved robots $\mathcal{A}_{\mathrm{res}}$ that are currently waiting at packing stations (Line~1-2).
The delivery station of the order remains fixed to the original station $S_k$.


For the bound robot $a_i$, Cooperative-TP constructs a baseline estimate assuming no cooperation. 
A tentative collision-free path $P_i^{\mathrm{base}}$ is computed using the underlying grid-based planner to complete both the remaining items $O_k^{\mathrm{rem}}$ and the newly added items $\Delta O_k(t_{\mathrm{upd}})$. 
The corresponding estimated remaining time is $
\mathrm{ETA}_i^{\mathrm{base}} = \frac{|P_i^{\mathrm{base}}|}{v},
$
where $v$ is the robot speed.
In parallel, a tentative path $P_i^{\mathrm{rem}}$ is computed assuming that $a_i$ only serves the remaining items $O_k^{\mathrm{rem}}$, with the corresponding estimated remaining time $ \mathrm{ETA}_i^{\mathrm{rem}} = \frac{|P_i^{\mathrm{rem}}|}{v} $
(Lines~5--6). 
For each reserved robot $a_j \in A_{\text{res}}$, Cooperative-TP constructs an additional pickup route $P_j^{\mathrm{add}}$ that starts from the current packing station of $a_j$, visits all pickup locations in $\Delta O_k(t)$, and delivers them to $S_k$, while avoiding reserved paths in the token. 
The corresponding estimated time is
$ \mathrm{ETA}_j^{\mathrm{add}} = \frac{|P_j^{\mathrm{add}}|}{v}$
(Lines~7--10).

\begin{figure*}[!t]
    \centering
    \setlength{\belowcaptionskip}{-0.5cm}
    \includegraphics[width=\linewidth]{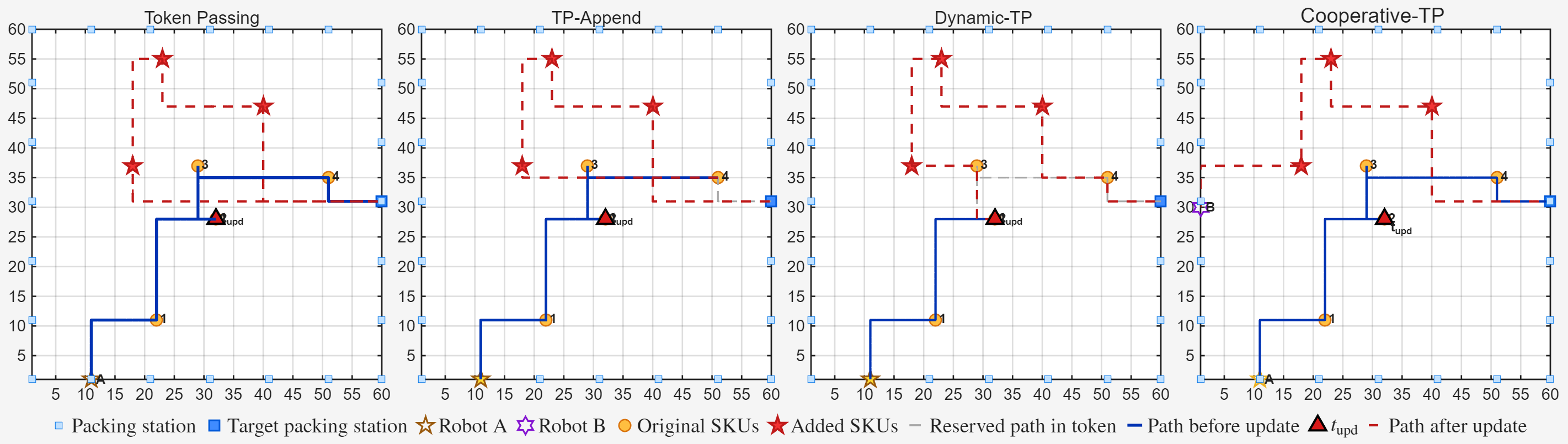}
    \caption{Representative case illustrating how different strategies handle a dynamic order update.}
    \label{case}
\end{figure*}

Cooperation is activated if there exists a reserved robot $a_{j^\star}$ such that the estimated completion time under cooperation
\begin{equation}
\max\left( \mathrm{ETA}_{j^\star}^{\mathrm{add}}, \mathrm{ETA}_i^{\mathrm{rem}} \right)
< \mathrm{ETA}_i^{\mathrm{base}}
\end{equation}
where the left-hand side represents the completion time under parallel execution of the two robots, determined by the slower one. 
When this condition is satisfied, the newly added SKUs $\Delta O_k(t_{\mathrm{upd}})$ are reassigned to $a_{j^\star}$ and removed from the task sequence of the bound robot $a_i$. 

When the cooperation condition is satisfied, the primary robot $a_i$ continues to serve the remaining SKUs $O_k^{\mathrm{rem}}$, while the newly added SKUs $\Delta O_k(t_{\mathrm{upd}})$ are assigned to the selected reserved robot $a_{j^\star}$. The selected robot is then inserted into the replanning set to compute its collision-free path using $\texttt{Path1}(\cdot)$ (Lines~11--15).

Moreover, both Dynamic-TP and Cooperative-TP preserve the deadlock-free 
execution and completeness properties of classical TP under the 
well-formed MAPD assumptions~\cite{ma2017lifelong}. Specifically, 
the number of orders and update events is finite, the shared token 
is accessed mutually exclusively, and every committed route is 
generated by the same reservation-based  $\texttt{Path1}(\cdot)$ / $\texttt{Path2}(\cdot)$
procedures used in classical TP. The proposed deadline-aware rule 
only determines the processing order of simultaneously triggered 
updates. It does not exclude any affected robot from token access, 
because all updates in the finite priority queue are processed 
sequentially. Hence, the proposed event-triggered mechanism does not 
introduce conflicting path reservations or starvation among updated 
orders. If $\texttt{Path1}(\cdot)$  temporarily fails, the robot moves to a 
non-task endpoint through $\texttt{Path2}(\cdot)$, and the unfinished order 
remains eligible for subsequent replanning. Therefore, every finite 
set of initially released and dynamically updated orders is eventually 
completed on a well-formed instance.
\subsection{Computational Complexity Analysis}
\textcolor{black}{Let $C_{\mathrm{plan}}$ denote the computational cost of one
single-robot path-planning call in the underlying MAPF planner. Each
invocation of $\texttt{Path1}(\cdot)$ or $\texttt{Path2}(\cdot)$ corresponds
to one A* search. When the time-expanded graph contains $N$ searchable
nodes, $C_{\mathrm{plan}}=\mathcal{O}(N\log N)$. For Dynamic-TP, a dynamic
order update first triggers a $\texttt{Path1}(\cdot)$ call for the affected
robot. If it fails, $\texttt{Path2}(\cdot)$ and subsequent replanning may
be required. Let $R_{\mathrm{upd}}^{(u)}$ denote the actual number of
planning calls associated with update event $u$. Its computational
complexity is therefore
$\mathcal{O}(R_{\mathrm{upd}}^{(u)}C_{\mathrm{plan}})$, and the total
complexity for $U$ update events is
$\mathcal{O}\!\left(\sum_{u=1}^{U}R_{\mathrm{upd}}^{(u)}
C_{\mathrm{plan}}\right)$.
In our simulations, the average number of planning calls per update is
$\bar R=1.04$, with only $4\%$ of update events requiring more than one
call and an observed maximum of two calls. Thus,
$R_{\mathrm{upd}}^{(u)}$ behaves as a small empirical constant in the
considered settings, and the observed average complexity per update
remains $\mathcal{O}(N\log N)$.}

\textcolor{black}{For Cooperative-TP, besides replanning the updated route of the bound
robot, the algorithm evaluates candidate pickup routes for reserved robots.
Let $n_{\mathrm{res}}^{(u)}$ denote the number of reserved robots at update
event $u$. Since a constant number of routes are evaluated for the bound
robot and one candidate route is evaluated for each reserved robot, the
computational complexity per update is
$\mathcal{O}((2+n_{\mathrm{res}}^{(u)})C_{\mathrm{plan}})$.
Accordingly, the total complexity for $U$ update events is
$\mathcal{O}\!\left(\sum_{u=1}^{U}
(2+n_{\mathrm{res}}^{(u)})C_{\mathrm{plan}}\right)$.}

\section{Simulation Results}

We evaluate the proposed methods in a grid-based RCWS environment that follows the RubikCell structure. Simulations are conducted on two cell sizes ($60\times60$ and $40\times80$) to study the influence of spatial scale. For each parameter setting, we generate 500 random instances and run all four strategies on the same instances for fair comparison. Reported values are averages over all trials. Dynamic parameters $(p_{\text{add}}, n_{\text{add}})$, and system size $N$ are varied across experiments to study dynamic intensity, scalability, and the effect of cooperative behavior. \textcolor{black}{We develop a simulation framework in MATLAB R2025b, which faithfully models the dynamics of multi-SKU order generation and grid-based motion planning within RubikCells. The simulation is executed on a laptop with an AMD Ryzen 7 5800H processor (3.20~GHz) and 16~GB RAM.}

All initial SKUs are randomly placed at storage locations. The initial number of each order is $3$. Robots begin at any packing station and are assigned to their initial orders by a static token-passing initialization.
For each experimental setting, we vary $p_{\text{add}}$ or \textcolor{black}{$n_{\mathrm{add}}$} while keeping other parameters fixed. Unless otherwise stated, active orders and primary robots are initialized in a one-to-one manner, with 10 additional robots reserved for cooperative assistance in Coop-TP. We evaluate four task planning strategies in the simulation study. Among them, Dynamic-TP and Cooperative-TP (Coop-TP) are proposed in Section~IV, while TP and TP-Append (TP-A) are included as baseline methods.

\begin{itemize}
    \item \textbf{Token Passing:} When new SKUs appear, they are treated as a new order that is permanently assigned to the same robot. After completing its original order and returning to the packing station, the robot starts a new TP cycle to serve the appended SKUs.
    
    \item  \textcolor{black}{\textbf{TP-Append:} A naive dynamic-handling strategy. The robot first follows its original route to collect all initially assigned SKUs. After completing these pickups, it replans a new route from the location of the last originally assigned SKU to retrieve the appended SKUs.}
\end{itemize}

\subsection{Representative Case Study}

We first present a representative case to illustrate how the four strategies handle a dynamic order update.
As shown in Fig.~3, a robot 'A' executes an order $O_k$ in a $60\times60$ RubikCell.
At time $t_{\mathrm{upd}}$, marked by the triangular flag, a set of new SKUs
$\Delta O_k(t_{\mathrm{upd}})$ is appended to the ongoing order.

Under TP, robot 'A' continues along its original reserved path and processes the added SKUs only after completing the original order, resulting in a long detour.
TP-Append improves this behavior by appending the new SKUs to the current task sequence, but replanning is still performed only after all original SKUs have been collected.

In contrast, Dynamic-TP replans immediately at $t_{\mathrm{upd}}$ by jointly considering the remaining original and newly added SKUs.
This allows robot 'A' to insert nearby added SKUs into the current route, reducing unnecessary backtracking compared with TP-A.
Cooperative-TP further exploits available reserved resources.
While robot 'A' continues serving the remaining original SKUs, a reserved robot 'B' is assigned to collect the newly added SKUs and deliver them to the same target packing station.
By parallelizing the workload, Coop-TP further reduces the completion time of the updated order compared with the single-robot strategies.

\subsection{Performance under Varying Order Dynamics}

\begin{figure}[!t]
    \centering
    \setlength{\belowcaptionskip}{-0.3cm}
    \begin{subfigure}{0.49\linewidth}
        \centering
        \includegraphics[width=\linewidth]{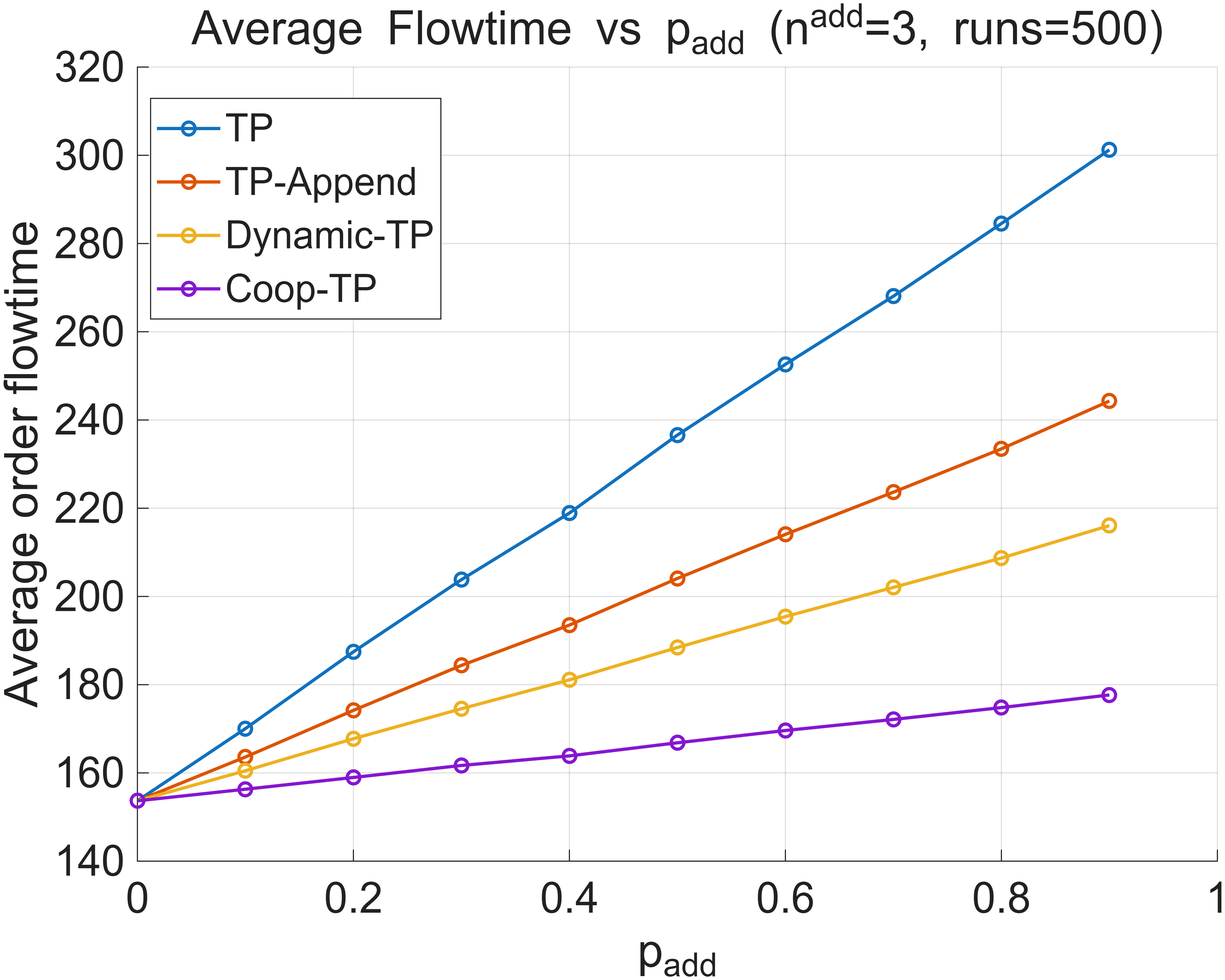}
        \caption{Impact of $p_{\mathrm{add}}$}
        \label{fig:p_add}
    \end{subfigure}
    \hfill
    \begin{subfigure}{0.49\linewidth}
        \centering
        \includegraphics[width=\linewidth]{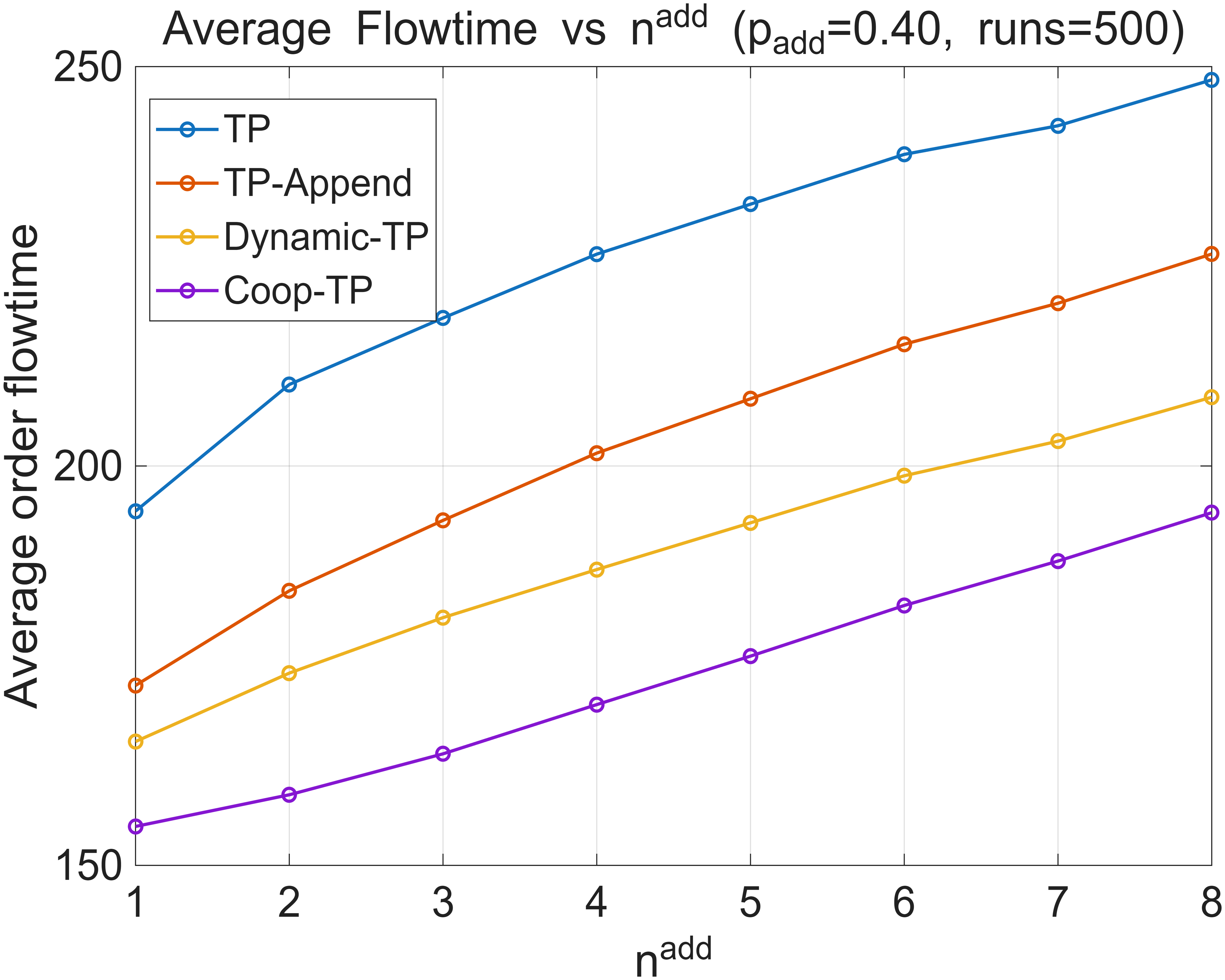}
        \caption{Impact of \textcolor{black}{$n_{\mathrm{add}}$}}
        \label{fig:n_add}
    \end{subfigure}
    \caption{\textcolor{black}{Overview of dynamic order fulfillment performance.}}
    \label{fig:dynamic_overview}
\end{figure}

\begin{table}[!t]
\centering
\caption{Average flowtime reduction (\%) relative to conventional TP under different order update probabilities $p_{\mathrm{add}}$.}
\label{tab:flowtime_reduction}
\setlength{\tabcolsep}{3.2pt}
\renewcommand{\arraystretch}{1.08}

\textcolor{black}{\begin{tabular}{c ccc ccc}
\toprule
\multirow{2}{*}{$p_{\mathrm{add}}$}
& \multicolumn{3}{c}{$60\times60$}
& \multicolumn{3}{c}{$40\times80$} \\
\cmidrule(lr){2-4}\cmidrule(lr){5-7}
& TP-A & Dynamic-TP & Coop-TP
& TP-A & Dynamic-TP & Coop-TP \\
\midrule
0.1 & 3.62\%  & 5.64\%  & 8.29\%
    & 3.69\%  & 5.60\%  & 8.13\% \\
0.3 & 8.60\%  & 13.22\% & 19.45\%
    & 9.05\%  & 13.72\% & 19.63\% \\
0.5 & 12.42\% & 18.98\% & 27.80\%
    & 12.90\% & 19.94\% & 28.19\% \\
0.7 & 15.48\% & 23.76\% & 34.39\%
    & 16.05\% & 24.28\% & 34.69\% \\
0.9 & 17.84\% & 26.95\% & 39.15\%
    & 18.09\% & 27.62\% & 39.77\% \\
\bottomrule
\end{tabular}}
\end{table}

Fig.~4 illustrates the average order flowtime under varying order-update
conditions. As shown in Fig.~4(a), when $n_{\mathrm{add}}=3$, increasing
$p_{\mathrm{add}}$ leads to a substantial increase in flowtime, particularly
for the conventional TP. On the $60\times60$ layout, the flowtime of TP
increases from $162.13$ at $p_{\mathrm{add}}=0.1$ to $284.72$ at
$p_{\mathrm{add}}=0.9$. In comparison, Dynamic-TP increases from $151.73$ to
$206.79$, while Coop-TP increases only from $146.78$ to $169.22$.
This indicates that the proposed methods are less sensitive to increasingly
frequent order updates. Fig.~4(b) shows a similar trend with respect to
$n_{\mathrm{add}}$, where larger SKU additions increase the fulfillment
burden for all strategies.

The performance differences can also be observed from representative
simulation results. At $(p_{\mathrm{add}},n_{\mathrm{add}})=(0.5,3)$ on
the $60\times60$ layout, the average flowtime is $223.80$, $194.04$,
$178.22$, and $156.17$ for TP, TP-A, Dynamic-TP, and Coop-TP, respectively.
TP-A provides an initial improvement by appending newly added SKUs after
the original pickup sequence. Dynamic-TP further reduces flowtime through
event-triggered replanning from the robot's current execution state, while
Coop-TP achieves the lowest flowtime through cooperative execution using
reserved robots.

\textcolor{black}{Table~\ref{tab:flowtime_reduction} further quantifies the relative performance
improvement by reporting the average flowtime reduction over all tested $n_{\mathrm{add}}$ values. The benefit of dynamic coordination becomes
increasingly pronounced as $p_{\mathrm{add}}$ increases. For the $60\times60$ layout, the average reduction achieved by Dynamic-TP increases from $5.64\%$ at $p_{\mathrm{add}}=0.1$ to $26.95\%$ at $p_{\mathrm{add}}=0.9$, while that of Coop-TP increases from $8.29\%$ to $39.15\%$. A consistent trend is observed for the $40\times80$ layout, where the corresponding reductions reach $27.62\%$ and $39.77\%$, respectively, at $p_{\mathrm{add}}=0.9$.}
\textcolor{black}{These results indicate that event-triggered replanning becomes increasingly
valuable as order updates become more frequent. Moreover, Coop-TP provides additional flowtime
reduction by exploiting reserved robotic resources for cooperative execution.}

\subsection{\textcolor{black}{Trade-off Between Dynamic-TP and Cooperative-TP}}

\begin{figure}[!t]
    \centering
    \setlength{\belowcaptionskip}{0cm}
    \includegraphics[width=\linewidth]{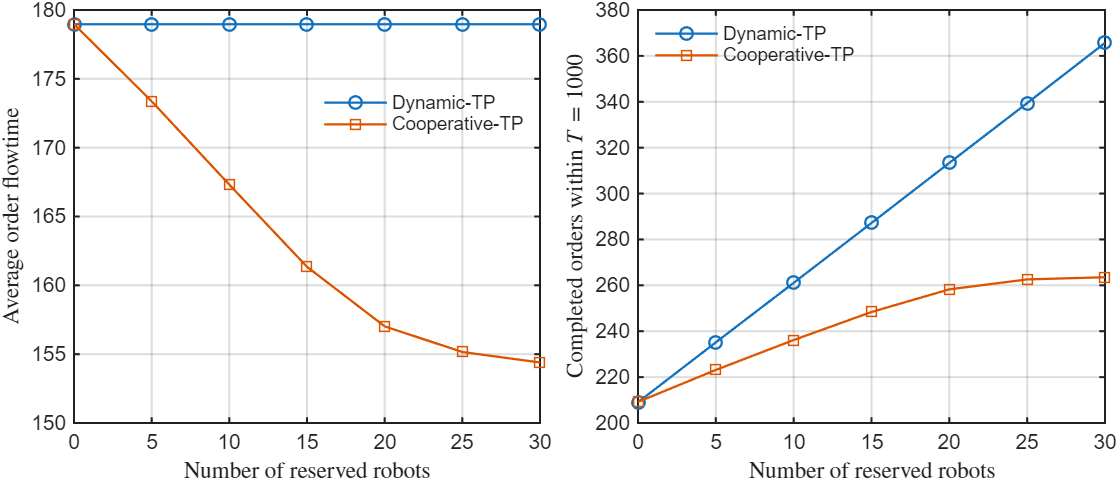}
 \caption{\textcolor{black}{Robot-resource trade-off in terms of average order flowtime and completed orders.}}
    \label{tradeoff}
\end{figure}

\textcolor{black}{Although Cooperative-TP can reduce the flowtime of dynamically updated orders
through parallel execution, such improvement requires additional robotic
resources. In contrast, Dynamic-TP preserves the one-order-to-one-robot binding
mode and can allocate available robots to independently serve more orders.
Therefore, the two execution modes exhibit an inherent trade-off between
individual-order fulfillment efficiency and system-level robot-resource
utilization.}

\textcolor{black}{To investigate this trade-off, we consider a $60\times60$ RubikCell with
40 primary robots as the basic configuration. The additional robot capacity is
varied as
$N_{\mathrm{res}}\in\{0,5,10,15,20,25,30\}$.
For Coop-TP, these robots are reserved for cooperative assistance when dynamic SKU
additions occur. For Dynamic-TP, the same additional robots directly join the
order-fulfillment queue and independently execute new orders.
Each order initially contains three SKUs, with
$p_{\mathrm{add}}=0.5$ and $n_{\mathrm{add}}=3$.
Two complementary experiments are conducted. First, 40 active orders are fixed
to evaluate how additional cooperative resources affect average order flowtime.
Second, a continuous backlog of 500 orders is maintained over a fixed simulation
horizon of $T=1000$ to evaluate the number of completed orders under the same
total robot capacity.}

\textcolor{black}{As shown in the left panel of Fig.~\ref{tradeoff}, the average flowtime of Dynamic-TP remains approximately $178.97$ because the additional robots are not used to
assist ongoing orders. In contrast, the flowtime of Coop-TP decreases from $178.97$
with no reserved robot to $154.40$ with 30 reserved robots, corresponding to a
$13.73\%$ reduction. The improvement gradually saturates as
$N_{\mathrm{res}}$ increases. The flowtime gain reaches $12.26\%$ with
20 reserved robots and increases only to $13.73\%$ with 30 robots.
This result shows that allocating more robots to cooperative execution improves
dynamic order handling, but with diminishing marginal benefit.}

\textcolor{black}{The right panel of Fig.~\ref{tradeoff} reveals the opposite effect from the
system-level fulfillment perspective. Within the same simulation horizon,
Dynamic-TP increases the average number of completed orders from $209.06$ with
40 robots to $365.66$ with 70 robots, since all additional robots can
independently process orders. For Coop-TP, the corresponding value increases only
from $209.19$ to $263.53$, because the additional robots are reserved for
cooperative assistance rather than independent order execution. Consequently,
the order-processing advantage of Dynamic-TP increases from $5.39\%$ at
$N_{\mathrm{res}}=5$ to $38.75\%$ at $N_{\mathrm{res}}=30$. This result highlights the higher efficiency of Dynamic-TP in terms of overall order-processing capacity.}

\textcolor{black}{These results reveal that the lower flowtime achieved by Coop-TP is not obtained
without resource cost. Reserving more robots for cooperation improves the
completion efficiency of dynamically updated orders, but reduces the robotic
capacity available for independently serving other orders. Dynamic-TP, although less
effective in minimizing the flowtime of individual updated orders, maintains
effective dynamic-order handling while achieving higher overall order-processing
capacity. Therefore, neither execution mode universally dominates the other.
The preferred strategy depends on robot-resource availability and operational
priority. Dynamic-TP is more suitable when robot resources are constrained or higher
system throughput is required, whereas Coop-TP is preferable when sufficient
robotic resources are available and shorter order flowtime is prioritized.
This trade-off provides a system-level interpretation of the two methods and offers practical guidance for robot-resource allocation in RCWS.}

Table~\ref{tab:runtime} reports the average online runtime per update event on a $60\times60$ grid with 40 active orders, $p_{\mathrm{add}}=0.5$, and $n_{\mathrm{add}}=3$. The number of robots varies from 45 to 70 to provide different numbers of reserved robots for cooperative execution.
Dynamic-TP maintains a stable runtime of approximately $20$--$27$ ms because replanning is localized to the affected robot. In contrast, Cooperative-TP increases from $92.06$ ms with 45 robots to $639.58$ ms with 70 robots, since additional reserved robots introduce more candidate routes to be evaluated during cooperative selection.

\begin{table}[!t]
\centering
\caption{\textcolor{black}{Average online runtime per update event (ms).}}
\label{tab:runtime}
\begin{tabular}{c|c|c}
\hline
\textbf{Robots} & \textbf{Dynamic-TP (ms)} & \textbf{Cooperative-TP (ms)} \\
\hline
45 & 21.74 & 92.06 \\
50 & 20.53 & 170.34 \\
55 & 26.81 & 326.05 \\
60 & 25.46 & 430.43 \\
65 & 25.86 & 525.45 \\
70 & 24.28 & 639.58 \\
\hline
\end{tabular}
\end{table}

The results therefore reveal a trade-off between computational overhead and coordination flexibility. Nevertheless, the runtime remains below $0.64$ s in all tested configurations, supporting the online applicability of both methods.





\section{Conclusion}
This paper introduces the Dynamic-MAPD problem to explicitly model online task replanning under dynamically evolving orders in RCWS. To address this problem, we develop an event-triggered token passing framework that enables timely and consistent replanning upon order updates. Within this framework, Dynamic-TP allows robots to update their routes directly from current positions, while a cooperative extension further leverages reserved robots to handle newly added items. \textcolor{black}{Future work will investigate richer dynamic order patterns, real-world implementations, and more flexible SKU-level task assignment strategies beyond the initial one-order-to-one-robot assignment paradigm.}

 \bibliographystyle{IEEEtran}
 \bibliography{Logistics}
\end{document}